\documentclass[10pt,twocolumn,letterpaper]{article}

\usepackage{cvpr}              

\usepackage[dvipsnames, svgnames, x11names]{xcolor}
\usepackage{color}
\usepackage{booktabs}
\usepackage{makecell}
\usepackage{multirow}
\usepackage{adjustbox}
\usepackage{caption}
\usepackage{subcaption}
\usepackage{booktabs}
\usepackage{pifont}
\newcommand{\gray}[1]{\textcolor{gray}{#1}}

\usepackage[accsupp]{axessibility}  

\definecolor{cvprblue}{rgb}{0.21,0.49,0.74}
\usepackage[pagebackref,breaklinks,colorlinks,allcolors=cvprblue]{hyperref}

\def\paperID{XXXX} 
\def\confName{CVPR}
\def\confYear{2026}

\title{SynCLIP: Synonym-Coherent Language-Image Pretraining for Robust Open-Vocabulary Dense Perception}

\author{Mingjie Xie$^{1}$\quad
Guangjun He$^{2}$\thanks{Corresponding author.}\quad
Dongli Xu$^{3}$\quad
Youtian Lin$^{4}$\quad
Hongjue Li$^{1}$\\
Pengming Feng$^{2}$\quad
Jian Guan$^{5}$\quad
Yue Deng$^{1,6}$\\
\small{$^{1}$Beihang University}\quad
\small{$^{2}$State Key Laboratory of Space Information System and Integrated Application}\\
\small{$^{3}$Independent Researcher}\quad
\small{$^{4}$Nanjing University}\quad
\small{$^{5}$Harbin Engineering University}\quad
\small{$^{6}$Beijing Zhongguancun Academy}
}

\begin{document}
\maketitle
\begin{abstract}
Open-vocabulary dense perception (OVDP) aims to localize objects unseen during training by leveraging textual knowledge. Despite the remarkable progress of recent CLIP-based approaches, we identify a critical limitation: synonym-induced grounding inconsistency, where semantically equivalent expressions yield disparate spatial attention patterns. This inconsistency undermines the robustness and performance of existing methods in real-world OVDP applications. To address this issue,  we propose SynCLIP, a Synonym-Coherent Language-Image Pretraining framework that enhances synonym-robust grounding for OVDP. SynCLIP introduces a Semantic-consistent Spatial Attention alignment (SSA) module to enhance spatial attention consistency by minimizing discrepancies between attention maps of original and synonymous expressions. Furthermore, a Spatial Attention Refinement (SAR) module selectively strengthens the most semantically relevant spatial regions within aligned maps for more precise and stable grounding. To support synonym-coherent pretraining, we also construct a Synonym-Enriched Visual Corpus (SEViC), which augments each category with multiple synonyms and textual definitions. Extensive experiments on multiple benchmarks demonstrate that SynCLIP substantially improves grounding consistency under diverse linguistic variants and achieves state-of-the-art performance among CLIP-based OVDP methods. Code is available at \url{https://github.com/Justlovesmile/SynCLIP}.
\end{abstract}    
\section{Introduction}
\label{sec:intro}
Open-vocabulary dense perception (OVDP) aims to recognize and localize objects from novel categories that are not predefined in the training set by leveraging textual knowledge \cite{zhu2024survey}. Unlike traditional dense perception tasks such as object detection and segmentation \cite{he2017mask, guan2023earl}, which operate within a fixed label space and thus struggle to generalize to real-world environments with an open set of object categories \cite{wu2023cora, wu2024towards}, OVDP employs textual expressions to flexibly represent labels. This language-driven supervision enables OVDP models to generalize beyond the closed-set categories when handling unseen categories. Consequently, OVDP has attracted growing attention for deployment in real-world applications such as robotics and autonomous driving \cite{zheng2024gaussiangrasper, zhou2024algpt}.

\begin{figure}[t]
    \centering
    \includegraphics[width=0.96\linewidth]{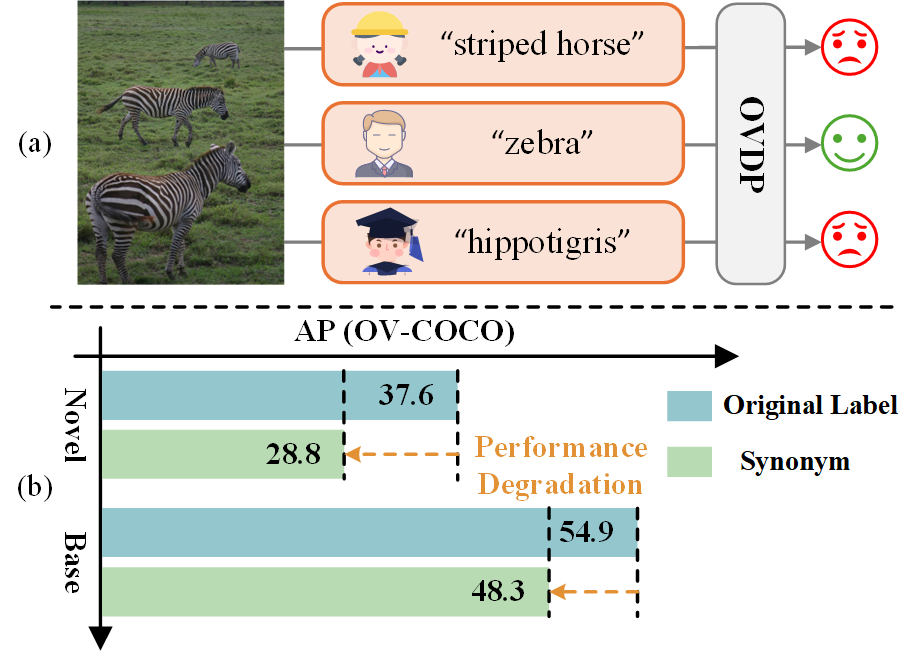}
    \caption{Illustration of synonym-induced grounding inconsistency. (a) shows inconsistent dense perception across diverse synonymous expressions. (b) presents performance degradation caused by synonym-induced grounding inconsistency. Here, `Novel' and `Base' denote the performance on unseen and seen categories, respectively.}
    \label{fig:question1}
    \vspace{-.46cm}
\end{figure}

\begin{figure}[t]
    \centering
    \includegraphics[width=0.96\linewidth]{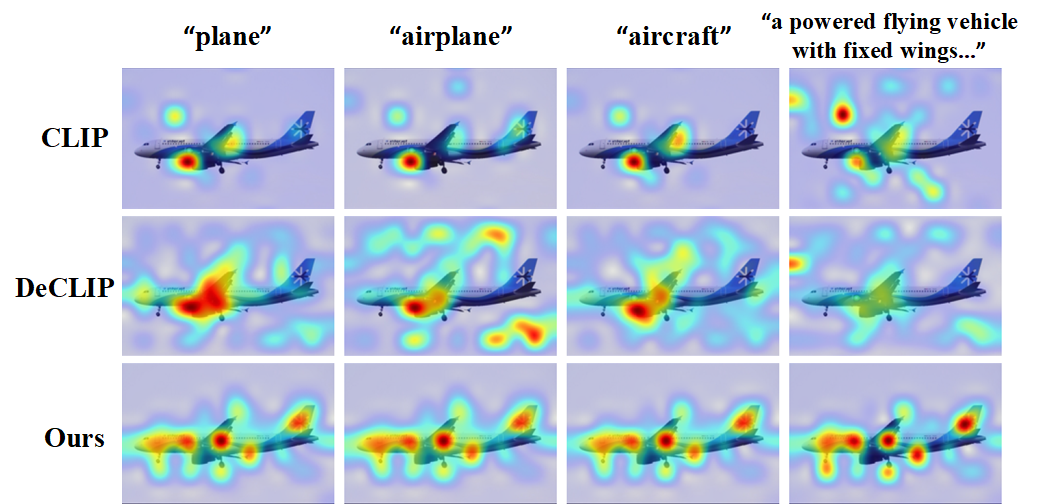}
    \caption{Comparison of spatial attention distributions generated by different models for synonymous expressions, \textit{e.g.}, synonyms and definition. Our method yields more consistent and semantically aligned attention, indicating synonym-robust grounding.}
    \label{fig:question}
    \vspace{-.46cm}
\end{figure}

To enable open-vocabulary capability, recent studies \cite{zhong2022regionclip, yang2025resclip, fu2025hierarchical} commonly leverage vision-language models (VLMs) pretrained on large-scale image–text pairs, such as CLIP \cite{radford2021learning, sun2023eva, li2023scaling}. These methods transfer the global vision–language alignment learned by VLMs to region-level representations, allowing visual regions to be associated with category-related textual expressions and thereby facilitating the recognition and localization of unseen categories. 
For example, a series of approaches \cite{wu2024clipself, wu2024clim, wang2025declip} enhance region-level vision–language alignment by employing techniques such as pseudo-region self-distillation or guidance from vision foundation models (VFMs). These strategies have led to notable improvements in OVDP tasks, including open-vocabulary detection (OVD) and segmentation (OVS).

Despite these advances, existing CLIP-based OVDP methods overlook a critical issue, \textit{i.e.,}  \textit{synonym-induced grounding inconsistency}. When diverse synonyms or semantically equivalent expressions are used to describe the same object, they often yield inconsistent spatial attention distributions. Such variations reduce model robustness and cause performance degradation in real-world OVDP applications. As shown in Figure~\ref{fig:question1}, simply replacing the original label (\textit{e.g.}, \textit{``zebra"}) with synonymous expressions (\textit{e.g.}, \textit{``striped horse"} and \textit{``hippotigris"}) results in a substantial drop in Average Precision (AP) for F-ViT \cite{wu2024clipself} on the OV-COCO benchmark \cite{lin2014microsoft}. This observation reveals the severity of synonym-induced grounding inconsistency and underscores the necessity for a synonym-coherent approach for real-world applications.

Figure~\ref{fig:question} further illustrates this issue through qualitative comparisons of several CLIP-based methods \cite{radford2021learning, wang2025declip}.  For the same image, different synonymous terms  (\textit{i.e.}, \textit{``plane"}, \textit{``airplane"} and \textit{``aircraft"}) and semantically equivalent expressions  (\textit{i.e.}, \textit{``a powered flying vehicle with fixed wings ..."}) yield markedly different spatial attention distributions, despite referring to the same visual object. Consequently, the model is guided toward different visual regions within the same image, as shown in rows one and two of Figure \ref{fig:question}. Although CLIP \cite{radford2021learning} demonstrates strong global representation capability, it lacks effective local region representation and may direct attention to irrelevant areas. Region-level alignment approaches, such as DeCLIP \cite{wang2025declip}, partially mitigate this limitation by reinforcing region-level vision–language alignment. However, they still fail to ensure consistent attention across synonymous expressions, ultimately leading to inconsistent localization.

To address this issue, we propose Synonym-Coherent Language-Image Pretraining (SynCLIP), a novel framework designed to achieve synonym-robust grounding for OVDP tasks. SynCLIP consists of two key components: a Semantic-consistent Spatial Attention alignment (SSA) module and a Spatial Attention Refinement (SAR) module, which jointly enforce consistent and precise spatial attention across synonymous expressions. Specifically, SSA enforces spatial attention consistency by minimizing the discrepancies between attention maps generated from original and semantically enriched synonymous expressions through a semantic alignment loss. This alignment effectively mitigates attention shifts caused by linguistic variation, thereby improving the robustness of dense perception. To further refine spatial grounding, SAR adopts a semantic token selection mechanism to identify the most semantically relevant spatial regions from enriched attention maps. It then leverages the spatial contextual reasoning capability of VFMs~\cite{oquab2023dinov2} to produce refined context-aware attention maps, which are subsequently aggregated to enhance attention accuracy and localization precision. In addition, we construct a Synonym-Enriched Visual Corpus (SEViC) that augments each category with multiple synonyms and textual definitions, serving as the foundation for synonym-coherent pretraining in SynCLIP.

The effectiveness of SynCLIP is validated on multiple OVDP benchmarks. Experimental results demonstrate that SynCLIP significantly improves grounding consistency across diverse synonymous expressions and achieves state-of-the-art performance among CLIP-based methods. The main contributions of this paper are summarized as follows:
\begin{itemize}
    \item We are the first to identify and analyze \textbf{synonym-induced grounding inconsistency} in CLIP-based methods, and reveal its negative impact on OVDP tasks.
    \item We propose SynCLIP, a novel framework that enhances \textbf{spatial attention consistency} across \textbf{semantically equivalent expressions}, thereby enhancing grounding robustness in real-world OVDP scenarios.
    \item We construct SEViC, a synonym-enriched image–text corpus that enhances category supervision with synonyms and definitions, providing \textbf{a foundation for synonym-coherent language–image pretraining}.
    \item Extensive experiments on multiple benchmarks verify the effectiveness of SynCLIP, achieving state-of-the-art performance and improved grounding consistency.
\end{itemize}
\section{Related Work}

\subsection{Vision-Language Pretraining}

Vision-language pretraining has emerged as a powerful paradigm for advancing visual understanding, particularly in open-vocabulary scenarios \cite{zhu2024survey}. Among existing methods, contrastive vision-language models, such as CLIP \cite{radford2021learning, sun2023eva}, ALIGN \cite{jia2021scaling} and Florence \cite{yuan2021florence}, have shown strong generalization and visual-language alignment by learning from web-scale image-text pairs. These models align image and text embeddings in a shared feature space through contrastive learning and have achieved remarkable success in tasks such as image-text retrieval and classification \cite{gan2022vision, lan2025band}. 
However, most of them perform alignment at the image level, relying on global representations without explicitly modeling the correspondence between local regions and textual expressions \cite{zhong2022regionclip}. This design limits their effectiveness in OVDP tasks, where precise spatial grounding and text-aligned region-level representation are essential.

To bridge this gap, several recent efforts have explored region-level vision-language alignment by adapting or fine-tuning CLIP to better model local interactions. For instance, CLIPSelf \cite{wu2024clipself} aligns region-level features with their corresponding cropped image-level representations, without requiring explicit region-text annotations. CLIM \cite{wu2024clim} treats randomly sampled patches as pseudo regions and aligns them with corresponding text in a mosaicked image, enabling spatially localized vision-language learning without box annotations. In contrast, DeCLIP \cite{wang2025declip} transfers spatial correlation patterns from VFMs for contextual consistency. These approaches have collectively improved the region-level alignment capabilities of CLIP-based models, thereby advancing their performance in OVDP tasks.

Despite these efforts, existing methods still fail to ensure consistent spatial attention across synonymous expressions. This limitation motivates our work, which explicitly addresses the need for synonym-robust grounding ability in real-world applications.

\subsection{Open-Vocabulary Dense Perception}

OVDP has attracted increasing attention for its ability to localize arbitrary visual concepts defined by textual categories. It primarily includes two sub-tasks, \textit{i.e.}, OVD and OVS. In OVD, methods like ViLD \cite{gu2022open}, VLDet \cite{lin2023learning} and GLIP \cite{li2022grounded} utilize CLIP-derived text embeddings to represent categories and align them with region proposals. These approaches typically treat text as classifiers or attention queries. Similarly, in OVS, methods like ZSSeg \cite{xu2022simple} and OVSeg \cite{liang2023open} usually follow a two-stage pipeline that first generates class-agnostic segmentation masks and then classifies each mask using CLIP-derived text embeddings. More recent approaches, such as SAN \cite{xu2023side} and CAT-Seg \cite{cho2024cat}, directly leverage CLIP embeddings to predict segmentation masks in an end-to-end manner. While these approaches successfully extend CLIP to OVDP tasks, they overlook the critical issue of synonym-induced grounding inconsistency, which significantly reduces the robustness of existing methods in real-world applications.
\begin{figure*}[htbp]
    \centering
    \includegraphics[width=1.0\linewidth]{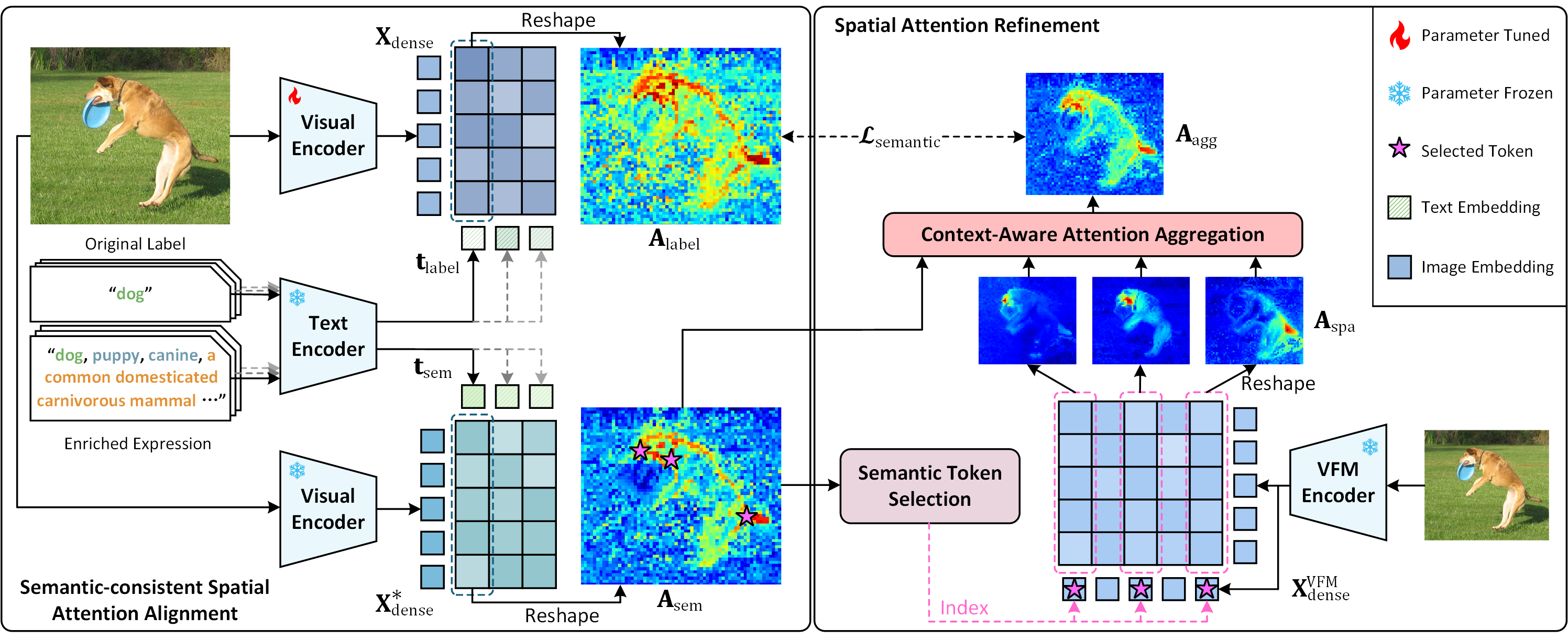}
    \caption{Illustration of the SynCLIP framework, which consists of two key components, \textit{i.e.}, semantic-consistent spatial attention alignment (SSA) and spatial attention refinement (SAR) modules. SSA enforces consistent grounding by aligning spatial attention between original labels and semantically enriched expressions, thereby reducing synonym-induced inconsistency. SAR refines attention precision by selecting the most semantically relevant tokens and aggregating context-aware spatial correlations derived from vision foundation models (VFMs). This synergistic design enables SynCLIP to achieve robust and precise grounding across diverse synonymous expressions.}
    \label{fig:architure}
    \vspace{-0.2cm}
\end{figure*}

\section{Proposed Method}
In this section, we introduce SynCLIP, a novel framework designed to enhance synonym-robust grounding ability for OVDP tasks. As illustrated in Figure \ref{fig:architure}, SynCLIP consists of two main components, \textit{i.e.}, semantic-consistent spatial attention alignment (SSA) and spatial attention refinement (SAR), which explicitly enforce consistent and precise spatial attention across diverse synonymous expressions.

\subsection{Preliminaries}
CLIP has become a foundational paradigm in vision-language research, which consists of a visual encoder $\mathcal{F}_{v}(\cdot)$ and a text encoder $\mathcal{F}_{t}(\cdot)$. The visual encoder is typically implemented using either convolutional neural networks (CNNs) \cite{he2016deep} or vision transformers (ViTs) \cite{dosovitskiy2020image}, while the text encoder usually adopts a standard transformer architecture to process textual inputs. In this study, we adopt the ViT-based variant of CLIP. Given an image-text pair, CLIP encodes the image $I\in\mathbb{R}^{3\times{h}\times{w}}$ and text $T$ into visual embedding $\mathbf{z}=\mathcal{F}_{v}(I)$ and text embedding $\mathbf{t}=\mathcal{F}_{t}(T)$, respectively. Here, $\mathbf{z}$ is the dense feature of the image, which is obtained by slightly modifying the last attention block of $\mathcal{F}_{v}(\cdot)$ following prior work \cite{zhou2022extract}. 

Specifically, let $\mathbf{x}=\{\mathbf{x}_0,\mathbf{x}_1,\cdots,\mathbf{x}_m\}$ denote the input tokens to the last attention block of visual encoder $\mathcal{F}_{v}(\cdot)$, where $\mathbf{x}\in\mathbb{R}^{(m+1)\times{d}}$ and $\mathbf{x}_0$ is the [CLS] token that captures the global representation of the image. The remaining tokens, each representing a $p \times p$ image patch, correspond to spatial locations, where $m=wh/p^2$ is the total number of image patches. Thus, $\mathbf{z}$ is calculated as follows:
\begin{align} &\mathbf{q}=\text{Proj}_q(\mathbf{x}),\mathbf{k}=\text{Proj}_k(\mathbf{x}),\mathbf{v}=\text{Proj}_v(\mathbf{x}),\\
&\mathbf{y}=\mathbf{x}+\text{Proj}(\mathbf{Attn}_{csa}\mathbf{v}),\\
&\mathbf{z}=\mathbf{y}+\text{FFN}(\mathbf{y}),
\end{align}
where $\mathbf{q}$, $\mathbf{k}$ and $\mathbf{v}$ represent the query, key and value embedding, respectively. $\text{Proj}(\cdot)$ and $\text{FFN}(\cdot)$ denote the projection layers and feed-forward network, respectively. $\mathbf{Attn}_{csa}$ represents the correlative self-attention process \cite{wang2024sclip}, which is formulated as follows:
\begin{align}
    \begin{aligned}
    \mathbf{Attn}_{csa}&=\text{SoftMax}(\mathbf{q}\mathbf{q}^\top/\tau)\\
    &+\text{SoftMax}(\mathbf{k}\mathbf{k}^\top/\tau),\\
    \end{aligned}
\end{align}
where the temperature coefficient $\tau$ is set to $\sqrt{d}$ by default. After obtaining the visual embedding $\mathbf{z}$, we discard the [CLS] token to attain dense feature representation $\mathbf{X}_{\text{dense}}\in\mathbb{R}^{m\times{d}}$. For ease of description, the following sections use $\mathbf{X}_{\text{dense}}=\mathcal{F}_v(I)$ to represent the above process.

\subsection{Semantic-consistent Spatial Attention Alignment}
To mitigate synonym-induced grounding inconsistency, we propose SSA, which promotes consistent spatial grounding by aligning the attention maps of original labels with those derived from semantically enriched expressions. 

As SSA relies on sufficient textual variations to support this alignment, we first construct a synonym-enriched visual corpus (SEViC), which augments category-level supervision with diverse synonymous expressions, including synonyms and definitions. SEViC serves as a foundational resource for learning robust language–image associations under expression diversity. Detailed construction of SEViC is described in Section \ref{section:sevic}.

Based on SEViC, SSA leverages a set of semantically enriched expressions as reference guidance. Specifically, given an input image $I$ and a set of original labels $\mathbb{T}_{\text{label}}=\{T_i^{\text{label}}\}_{i=1}^{n}$ (\textit{e.g.}, \textit{``dog"}), SSA first constructs the corresponding semantically enriched synonymous expressions $\mathbb{T}_{\text{sem}}=\{T_i^{\text{sem}}\}_{i=1}^{n}$ by incorporating synonyms and definitions (\textit{e.g.}, \textit{``dog, puppy, canine, a common domesticated carnivorous mammal ..."}). These enriched expressions provide more comprehensive semantic grounding cues, whose attention maps tend to localize object regions more accurately and consistently. Therefore, they serve as reliable references for aligning the spatial attention of original labels. 
After that, both $\mathbb{T}_{\text{label}}$ and $\mathbb{T}_{\text{sem}}$ are encoded by the frozen text encoder $\mathcal{F}_{t}(\cdot)$ of CLIP, yielding sentence-level text embeddings $\mathbf{t}_{\text{label}}\in\mathbb{R}^{n\times{d}}$ and $\mathbf{t}_{\text{sem}}\in\mathbb{R}^{n\times{d}}$, where $n$ denotes the number of expressions and $d$ is the embedding dimension. 

To model semantically consistent grounding, SSA employs a teacher-student dual visual encoder architecture to extract dense visual representations. Given an input image $I$, the student model $\mathcal{F}_v(\cdot)$ and its frozen counterpart $\mathcal{F}_v^{\ast}(\cdot)$, serving as the teacher model, generate dense feature representations $\mathbf{X}_{\text{dense}} = \mathcal{F}_v(I) \in \mathbb{R}^{m \times d}$ and $\mathbf{X}_{\text{dense}}^{\ast} = \mathcal{F}_v^{\ast}(I) \in \mathbb{R}^{m \times d}$, respectively. SSA then computes attention maps between visual features and both original and semantically enriched text embeddings, establishing a foundation for spatial attention alignment.
\begin{equation}
    \mathbf{A}_{\text{label}}=\frac{\mathbf{t}_{\text{label}}{\mathbf{X}_{\text{dense}}}^\top}{||\mathbf{t}_\text{label}||\cdot||\mathbf{X}_{\text{dense}}||},
\end{equation}
\begin{equation}
    \mathbf{A}_{\text{sem}}=\frac{\mathbf{t}_{\text{sem}}{\mathbf{X}^{\ast}_{\text{dense}}}^\top}{||\mathbf{t}_{\text{sem}}||\cdot||\mathbf{X}^{\ast}_{\text{dense}}||},
\end{equation}
where $\mathbf{A}\in\mathbb{R}^{n\times{m}}$ denotes the spatial attention maps between the text and image embeddings. To enforce semantic consistency across diverse expressions referring to the same concept, SSA introduces the semantic alignment loss:
\begin{equation}
    \mathcal{L}_{\text{semantic}}=\frac{1}{nm}\sum_{i=1}^{n}\sum_{j=1}^{m}||\mathbf{A}^{i,j}_{\text{label}}-\mathbf{A}^{i,j}_{\text{sem}}||_2.
\end{equation}

By minimizing the discrepancy between the attention distributions derived from original and enriched expressions, SSA encourages the model to produce consistent spatial attention across synonymous or semantically equivalent expressions, thereby enhancing the stability and coherence of spatial grounding regardless of expression diversity.

\subsection{Spatial Attention Refinement}

To further improve spatial grounding precision, we propose SAR, which consists of two stages, \textit{i.e.}, semantic token selection to identify the most semantically relevant spatial locations, and context-aware attention aggregation to refine spatial attention through contextual reasoning. SAR leverages the spatial representation capability of vision foundation models (VFMs)~\cite{oquab2023dinov2} to enhance both spatial accuracy and semantic coherence, leading to more reliable localization across synonymous expressions.

\subsubsection{Semantic Token Selection}
Given the attention maps of semantically enriched expressions $\mathbf{A}_{\text{sem}}$, which capture the correlation between dense feature representation $\mathbf{X}^{\ast}_{\text{dense}}$ and the enriched text embedding $\mathbf{t}_{\text{sem}}$, SAR first identifies the most semantically relevant spatial locations via a semantic token selection mechanism. Specifically, the top-$k$ tokens with the highest attention scores are selected for each enriched expression: 

\begin{equation}
\mathcal{K}=\text{TopK}(\mathbf{A}_{\text{sem}}, k),
\end{equation}
where $\mathcal{K}\in\mathbb{R}^{n\times{k}}$ stores the indices of selected tokens across $n$ text embeddings. A pretrained VFM model, denoted as $\mathcal{F}_v^{\text{VFM}}(\cdot)$, extracts the dense visual feature $\mathbf{X}^{\text{VFM}}_{\text{dense}}\in\mathbb{R}^{m\times{d}}$ that encodes rich spatial context. Here, $\mathbf{X}^{\text{VFM}}_{\text{dense}}=\{\mathbf{x}_1,\mathbf{x}_2,\cdots,\mathbf{x}_m\}$ represents the image tokens of VFM feature, which has the same shape of $\mathbf{X}^{\ast}_{\text{dense}}$.

Based on the indices in $\mathcal{K}$, SAR computes the spatial correlation matrix $\mathbf{S}$ by evaluating cosine similarity between each selected token and all image tokens:
\begin{equation}
    \mathbf{s}_{i,j}=\frac{\mathbf{x}_i\cdot\mathbf{x}_j}{\max(||\mathbf{x}_i||\cdot||\mathbf{x}_j||,\epsilon)},
\end{equation}
where $\epsilon$ is set to $10^{-8}$ by default to avoid division by zero at extrema. This results in a set of $k$ spatial correlation matrices per text embedding, \textit{i.e.}, $\mathbf{S}=\{\mathbf{s}_{i,j} | i\in\mathcal{K},j\in[1,m]\}\in\mathbb{R}^{n\times{k}\times{m}}$, each of which represents the spatial correlation from a key semantic location to the entire image.

\subsubsection{Context-Aware Attention Aggregation}

\begin{figure}[t]
    \centering
    \includegraphics[width=1.0\linewidth]{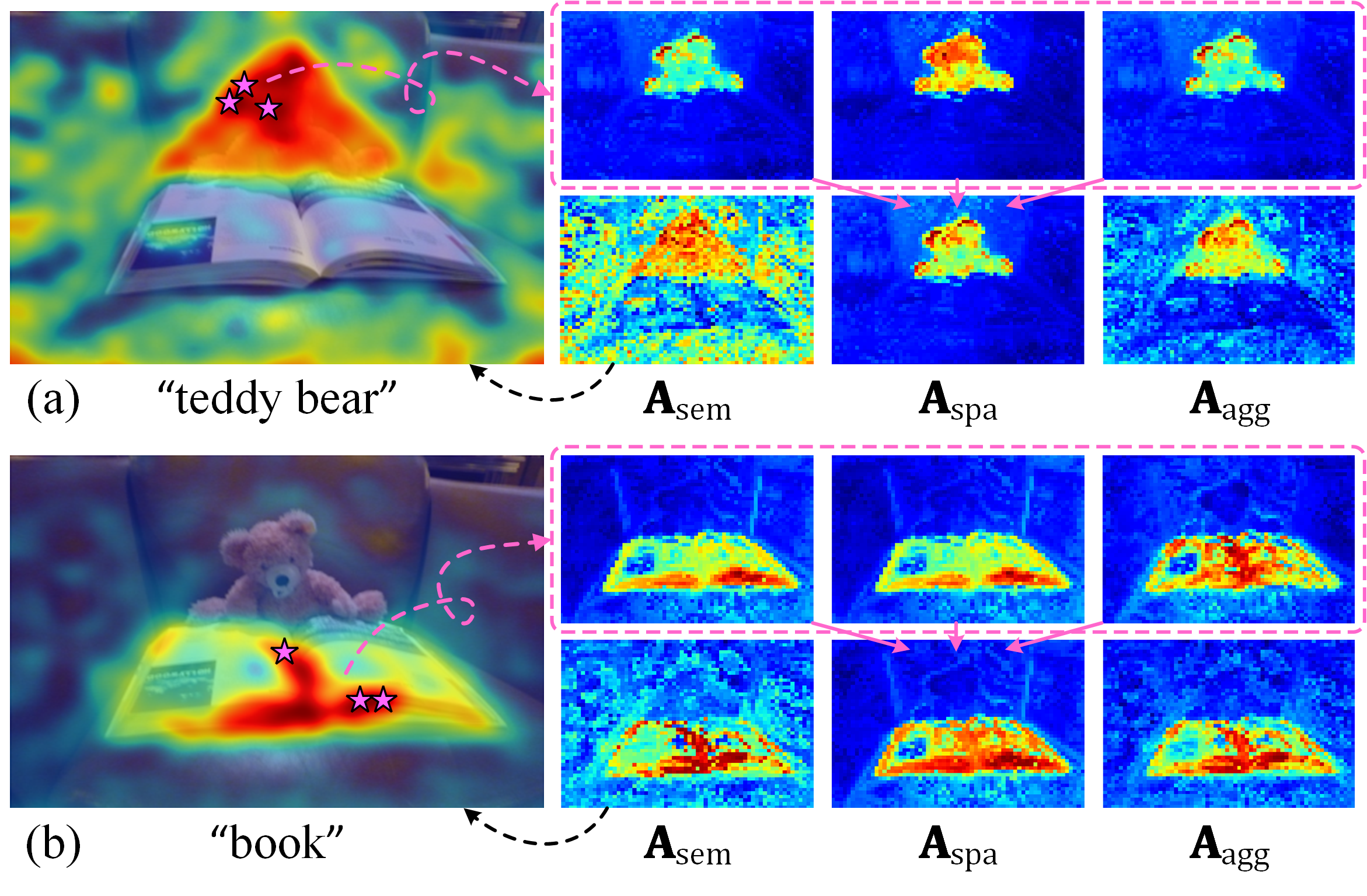}
    \caption{Illustration of the SAR process with $k{=}3$. From left to right, we visualize the semantic attention $\mathbf{A}_{\text{sem}}$, the spatial correlation attention $\mathbf{A}_{\text{spa}}$, and the aggregated attention $\mathbf{A}_{\text{agg}}$. The aggregated map preserves both semantic relevance and text-guided spatial localization. 
    }
    \label{fig:tts}
    \vspace{-0.1cm}
\end{figure}

To obtain a refined and semantically coherent attention map, the spatial correlation matrices $\mathbf{S}$ are aggregated into a final attention map $\mathbf{A}_{\text{spa}}=\text{Aggregator}(\mathbf{S})\in\mathbb{R}^{n\times m}$, where $\text{Aggregator}(\cdot)$ computes the mean across the $k$ selected tokens. This simple averaging strategy proves empirically effective, as it highlights spatial regions that are consistently correlated with the most informative semantic cues. Figure~\ref{fig:tts} visualizes this process, showing how semantic attention, spatial correlation, and aggregated attention jointly contribute to SAR.

Finally, to balance semantic relevance and spatial precision, SAR fuses the semantic attention $\mathbf{A}_{\text{sem}}$ and the spatial correlation attention $\mathbf{A}_{\text{spa}}$ as:
\begin{equation}
    \mathbf{A}_{\text{agg}} = \alpha \mathbf{A}_{\text{spa}} + \beta \mathbf{A}_{\text{sem}},
\end{equation}
where $\alpha$ and $\beta$ control the trade-off between spatial accuracy and semantic consistency, and the effect of their values is further analyzed in the Appendix. 
The refined attention $\mathbf{A}_{\text{agg}}$ supervises grounding accuracy and consistency by minimizing the optimized semantic alignment loss:
\begin{equation}
    \mathcal{L}^{+}_{\text{semantic}}=\frac{1}{nm}\sum_{i=1}^{n}\sum_{j=1}^{m}\|\mathbf{A}^{i,j}_{\text{label}}-\mathbf{A}^{i,j}_{\text{agg}}\|_2.
\end{equation}

Through this refinement and aggregation mechanism, SAR enhances spatial fidelity of attention maps while preserving the semantics conveyed by enriched textual expressions, thereby ensuring semantic coherence and accurate spatial grounding across diverse synonymous expressions for robust performance in real-world OVDP scenarios.
\section{Experiments}

\subsection{Datasets and Metrics}
\label{sec:datasets}

\subsubsection{Synonym-Enriched Visual Corpus}
\label{section:sevic}

\begin{table*}[t]
    \centering
    \begin{tabular}{cccc}
        \toprule
        \# & Class & Definition & Synonyms \\
        \midrule
        1 & boat & \textit{a small vessel for traveling on water ...} & \textit{vessel, watercraft, small boat, ...} \\
        2 & giraffe & \textit{tall animal having a very long neck and legs ...} & \textit{giraffa camelopardalis, long-necked animal, ...} \\
        3 & frisbee & \textit{a light, plastic disk propelled with a flip ...} & \textit{flying disc, throw disc, wham-o disc, ...} \\
        \bottomrule
    \end{tabular}
    \caption{Examples of synonym-enriched textual expressions in SEViC. For each category, SEViC includes a concise definition and multiple semantically consistent synonyms, which are used to enrich linguistic supervision during synonym-coherent pretraining.}
    \label{tab:sevic}
    \vspace{-0.3cm}
\end{table*}

To enable synonym-coherent language–image pretraining for robust OVDP, we construct a synonym-enriched visual corpus (SEViC). SEViC augments category-level supervision with semantically consistent expressions, including synonyms and textual definitions, encouraging consistent grounding across synonymous expressions.

To ensure generalizability, SEViC is built upon the training images of COCO2017 dataset \cite{lin2014microsoft}, a widely used benchmark offering diverse visual contexts. Since LVIS \cite{gupta2019lvis} extends COCO2017 with finer-grained and long-tailed annotations over the same image set, we adopt category names from both datasets to form a comprehensive vocabulary of 1,232 unique categories, covering both frequent and rare concepts. In addition, two meta-categories, \textit{``object"} and \textit{``background"}, are further added to represent special global semantics. To enrich linguistic supervision, we incorporate the LVIS-provided synonyms and definitions where available, and employ large language models (LLMs), \textit{e.g.}, DeepSeek \cite{liu2024deepseek}, to automatically generate multiple synonyms and concise definitions, as illustrated in Table \ref{tab:sevic}. All generated expressions are validated via LLM-based consistency checks to ensure semantic fidelity. Each image is finally paired with its associated categories and verified expressions, forming a synonym-enhanced image–text corpus comprising 118,287 images, 1,234 categories, and 11,558 semantically enriched expressions.

SEViC serves as the foundation for pretraining SynCLIP, enabling the model to learn synonym-robust dense perception, thereby mitigating the synonym-induced grounding inconsistency observed in CLIP-based models.

\subsubsection{Downstream Finetuning Datasets}
We conduct evaluations on two widely used benchmarks, \textit{i.e.}, OV-COCO \cite{lin2014microsoft} and OV-LVIS \cite{gupta2019lvis}. For OV-COCO, following the setup of previous work \cite{wu2024clim}, we divide the 80 classes in the COCO dataset \cite{lin2014microsoft} into 48 base classes and 17 novel classes. The training set comprises 107,761 images containing annotations for base classes only, while evaluation is performed on a test set of 4,836 images with annotations for both base and novel classes. Detection performance is measured using the box Average Precision (AP) at IoU threshold 0.5, denoted as $\text{AP}_{50}$, with a specific focus on $\text{AP}_{50}^{\text{novel}}$, which reflects the model's ability to generalize to novel classes. 

Regarding OV-LVIS, we retain 866 common and frequent classes in the LVIS dataset \cite{gupta2019lvis} as base classes and 337 rare classes as novel classes, following the setup in ViLD \cite{gu2022open}. Similarly, the training annotations are limited to base classes. We report mean Average Precision (mAP) of masks averaged over IoU thresholds from 0.5 to 0.95. The performance on rare (\textit{i.e.}, novel) classes, denoted as mAP$^{\text{mask}}_r$, is the primary evaluation metric in experiments.

\subsection{Implementation Details}
\label{sec:details}
SynCLIP is implemented based on the DeCLIP framework \cite{wang2025declip}, upon which we integrate the proposed SSA and SAR modules. The default weight of semantic alignment loss is set to 0.05. We use DINOv2 \cite{oquab2023dinov2} as the VFM encoder. The balance coefficients for attention aggregation, \textit{i.e.}, $\alpha$ and $\beta$, are both set to 0.5 by default. Training is conducted using 4 GPUs with a batch size of 8 per GPU. We train our SynCLIP for 6 epochs using the AdamW \cite{loshchilov2019decoupled} optimizer with a learning rate of $2\times10^{-5}$ and a weight decay of $0.1$. 

For downstream tasks, we adopt F-ViT \cite{wu2024clipself} as the baseline, forming our method, \textit{i.e.}, F-ViT+SynCLIP. All other hyper-parameters follow the default settings of the respective baselines. Following a previous study \cite{wu2024clipself, wang2025declip}, we ensure fair comparisons using the same training data as the corresponding baselines.

\begin{table*}[t]
    \label{tab:ovod}
    \centering
    \begin{subtable}[t]{.48\textwidth}
        \centering
        \begin{adjustbox}{width=\textwidth,center,valign=t}
        \begin{tabular}{l|c|c|c}
          \toprule
          Method & Venue & Backbone & AP$_{50}^{\text{novel}}$ \\
          \midrule
          ViLD \cite{gu2022open}              & ICLR2022 & RN50     & 27.6  \\
          F-VLM \cite{kuo2023f}               & ICLR2023 & RN50     & 28.0  \\
          OV-DETR \cite{zang2022open}         & ECCV2022 & RN50     & 29.4  \\
          BARON-KD \cite{wu2023aligning}      & CVPR2023 & RN50     & 34.0  \\
          RegionCLIP \cite{zhong2022regionclip} & CVPR2022 & RN50x4 & 39.3  \\
          CORA$^+$ \cite{wu2023cora}          & CVPR2023 & RN50x4   & 43.1  \\
          OV-DQUO \cite{wang2025ov}           & AAAI2025 & RN50x4   & 45.6  \\
          \midrule
          RO-ViT \cite{kim2023region}         & CVPR2023 & ViT-L/16 & 33.0  \\
          CFM-ViT \cite{kim2023contrastive}   & ICCV2023 & ViT-L/16 & 34.1  \\
          F-ViT+CLIPSelf \cite{wu2024clipself} & ICLR2024 & ViT-B/16 & 37.6  \\
          F-ViT+CLIPSelf \cite{wu2024clipself} & ICLR2024 & ViT-L/14 & 44.3  \\
          F-ViT+DeCLIP \cite{wang2025declip}   & CVPR2025 & ViT-B/16 & 41.1  \\
          F-ViT+DeCLIP \cite{wang2025declip}   & CVPR2025 & ViT-L/14 & 46.2  \\
          \midrule
          \textbf{F-ViT+SynCLIP (Ours)}        & CVPR2026 & ViT-B/16 & \textbf{43.6} \\
          \textbf{F-ViT+SynCLIP (Ours)}        & CVPR2026 & ViT-L/14 & \textbf{49.8} \\
        \bottomrule
        \end{tabular}
        \end{adjustbox}
        \caption{Results on OV-COCO benchmark}
        \label{tab:ovod_coco}
    \end{subtable}
    \hfill
    \begin{subtable}[t]{.494\textwidth}
        \begin{adjustbox}{width=\textwidth,center,valign=t}
        \begin{tabular}{l|c|c|c}
          \toprule
          Method & Venue & Backbone & mAP$^{\text{mask}}_{r}$ \\
          \midrule
          ViLD \cite{gu2022open}                & ICLR2022 & RN50     & 16.3  \\
          OV-DETR \cite{zang2022open}           & ECCV2022 & RN50     & 17.4  \\
          RegionCLIP \cite{zhong2022regionclip} & CVPR2022 & RN50x4   & 22.0  \\
          BARON-KD \cite{wu2023aligning}        & CVPR2023 & RN50     & 22.6  \\
          OV-SAM \cite{yuan2024open}            & CVPR2024 & RN50x16  & 24.0  \\
          CORA$^+$ \cite{wu2023cora}            & CVPR2023 & RN50x4   & 28.1  \\
          F-VLM \cite{kuo2023f}                 & ICLR2023 & RN50x64  & 32.8  \\
          \midrule
          PromptOVD \cite{song2023prompt}       & Arxiv2023 & ViT-B/16 & 23.1  \\
          RO-ViT \cite{kim2023region}           & CVPR2023 & ViT-L/16 & 32.4  \\
          F-ViT+CLIPSelf \cite{wu2024clipself}  & ICLR2024 & ViT-B/16 & 25.3  \\
          F-ViT+CLIPSelf \cite{wu2024clipself}  & ICLR2024 & ViT-L/14 & 34.9  \\
          F-ViT+DeCLIP \cite{wang2025declip}    & CVPR2025 & ViT-B/16 & 26.8  \\
          F-ViT+DeCLIP \cite{wang2025declip}    & CVPR2025 & ViT-L/14 & \textbf{37.2}  \\
          \midrule
          \textbf{F-ViT+SynCLIP (Ours)}         & CVPR2026 & ViT-B/16 & \textbf{27.8} \\
          \textbf{F-ViT+SynCLIP (Ours)}         & CVPR2026 & ViT-L/14 & \textbf{37.2} \\
          \bottomrule
        \end{tabular}
        \end{adjustbox}
        \caption{Results on OV-LVIS benchmark}
        \label{tab:ovod_lvis}
    \end{subtable}
    \caption{Comparison with state-of-the-art methods. `RN50' represents ResNet50 \cite{he2016deep}, where `RN50x4' is scaled up by a factor of 4 according to the scaling rules \cite{tan2019efficientnet}. `B' and `L' in ViT-based methods stand for base and large model sizes. `/16' and `/14' indicate the downsample ratio of input images. The best results among methods using comparable backbones are highlighted in bold.}
\end{table*}

\subsection{Main Results}
\label{sec:results}
We evaluate SynCLIP on two representative benchmarks \cite{lin2014microsoft, gupta2019lvis} 
to verify both its general performance and robustness to linguistic variations. Experiments are conducted in two stages. We first conduct benchmark comparisons with state-of-the-art methods to demonstrate the effectiveness of synonym-coherent pretraining. We then perform a synonym-robustness evaluation under synonym substitution, further validating SynCLIP's ability to maintain consistent grounding across diverse textual expressions.

\subsubsection{Benchmark Comparison}
To assess whether introducing synonym-coherent alignment benefits general OVDP performance, we evaluate SynCLIP under standard protocols. The core insight is that integrating synonym-consistent supervision during pretraining helps the model learn stable and semantically invariant vision–language attention. By reducing inconsistencies across synonymous expressions, SynCLIP acquires stronger category semantics, leading to improved novel-class recognition and generalization.

\noindent\textbf{Results on OV-COCO:} As shown in Table~\ref{tab:ovod_coco}, SynCLIP significantly enhances the performance over F-ViT on the OV-COCO benchmark, achieving 43.6 and 49.8 $\text{AP}_{50}^{\text{novel}}$ with ViT-B/16 and ViT-L/14 backbones, respectively. Beyond raw accuracy, the improvement highlights a more semantically stable vision–language representation, where different lexical forms referring to the same concept yield consistent region–text alignment. This stability directly translates into stronger recognition of unseen classes, confirming that SynCLIP’s synonym-coherent pretraining narrows the linguistic gap between base and novel categories, enabling SynCLIP to outperform previous methods on the OV-COCO.

\noindent\textbf{Results on OV-LVIS:} Table~\ref{tab:ovod_lvis} presents results on the more challenging OV-LVIS benchmark. FViT+SynCLIP achieves 27.8 and 37.2 mAP$^{\text{mask}}_r$ with ViT-B/16 and ViT-L/14 backbones, respectively. The former surpasses all compared methods with similar backbones, while the latter is comparable to the best existing result under the large-backbone setting.
It is worth noting that OV-LVIS poses a fundamentally more challenging scenario than OV-COCO, as it involves fine-grained categories with long-tailed distributions and extensive lexical overlap. Under such conditions, models tend to overfit to frequent names and struggle to generalize across semantically similar yet lexically diverse concepts. SynCLIP maintains competitive performance in this setting, indicating its ability to transfer category semantics beyond surface lexical cues.

Across both OV-COCO and OV-LVIS, the consistent improvements can be attributed to SynCLIP’s synonym-coherent pretraining, which provides more stable and semantically aligned region–text attention. SSA reduces attention variation caused by lexical differences, and SAR further refines spatial grounding with context-aware cues. These two components jointly strengthen the underlying vision–language alignment, allowing SynCLIP to capture clearer category semantics and generalize more effectively to novel categories. This explains why SynCLIP is able to outperform existing SOTA methods under standard open-vocabulary settings.

\subsubsection{Synonym-Robust Dense Perception}

\begin{table}[t]
    \centering
    \setlength{\tabcolsep}{1mm}
    \begin{tabular}{l|c|cc}
        \toprule
        Method & syn. & AP$_{50}^{\text{novel}}$ & \gray{AP$_{50}^{\text{base}}$} \\
        \midrule
        F-ViT+CLIPSelf \cite{wu2024clipself} & \ding{56} & 37.6        & \gray{54.9} \\
        F-ViT+CLIPSelf \cite{wu2024clipself} & \ding{52} & 28.8 (-8.8) & \gray{48.3 (-6.6)} \\
        \midrule
        F-ViT+DeCLIP$^{\dagger}$ \cite{wang2025declip} & \ding{56} & 41.0        & \gray{50.9} \\
        F-ViT+DeCLIP \cite{wang2025declip}             & \ding{52} & 31.5 (-9.5) & \gray{48.7 (-2.2)} \\
        \midrule
        F-ViT+SynCLIP (Ours) & \ding{56} & 43.6                 & \gray{51.8}        \\
        F-ViT+SynCLIP (Ours) & \ding{52} & \textbf{39.2 (-4.4)} & \textbf{\gray{51.2 (-0.6)}} \\
        \bottomrule
    \end{tabular}
    \caption{Robustness evaluation under synonym substitution on OV-COCO. \text{AP}$^{\text{novel}}_{50}$ measures grounding performance on novel classes, and \text{AP}$^{\text{base}}_{50}$ on base classes. `syn.' indicates evaluation where category names are replaced by synonymous expressions, revealing each model's sensitivity to linguistic variation. $\dagger$ denotes reproduced results.}
    \label{tab:synonyms_robust}
    \vspace{-0.3cm}
\end{table}

While the benchmark results demonstrate overall improvement, a key motivation of our SynCLIP is to enhance robustness when the original category names are replaced with synonymous expressions. To assess this, we evaluate grounding performance under a synonym substitution setting, following the same protocol used for comparing against state-of-the-art OVDP methods.

\begin{figure}[t]
    \centering
    \subfloat[Top: \textit{``banner"} vs. Bottom: \textit{``advertising banner"}]{\includegraphics[width=1.0\linewidth]{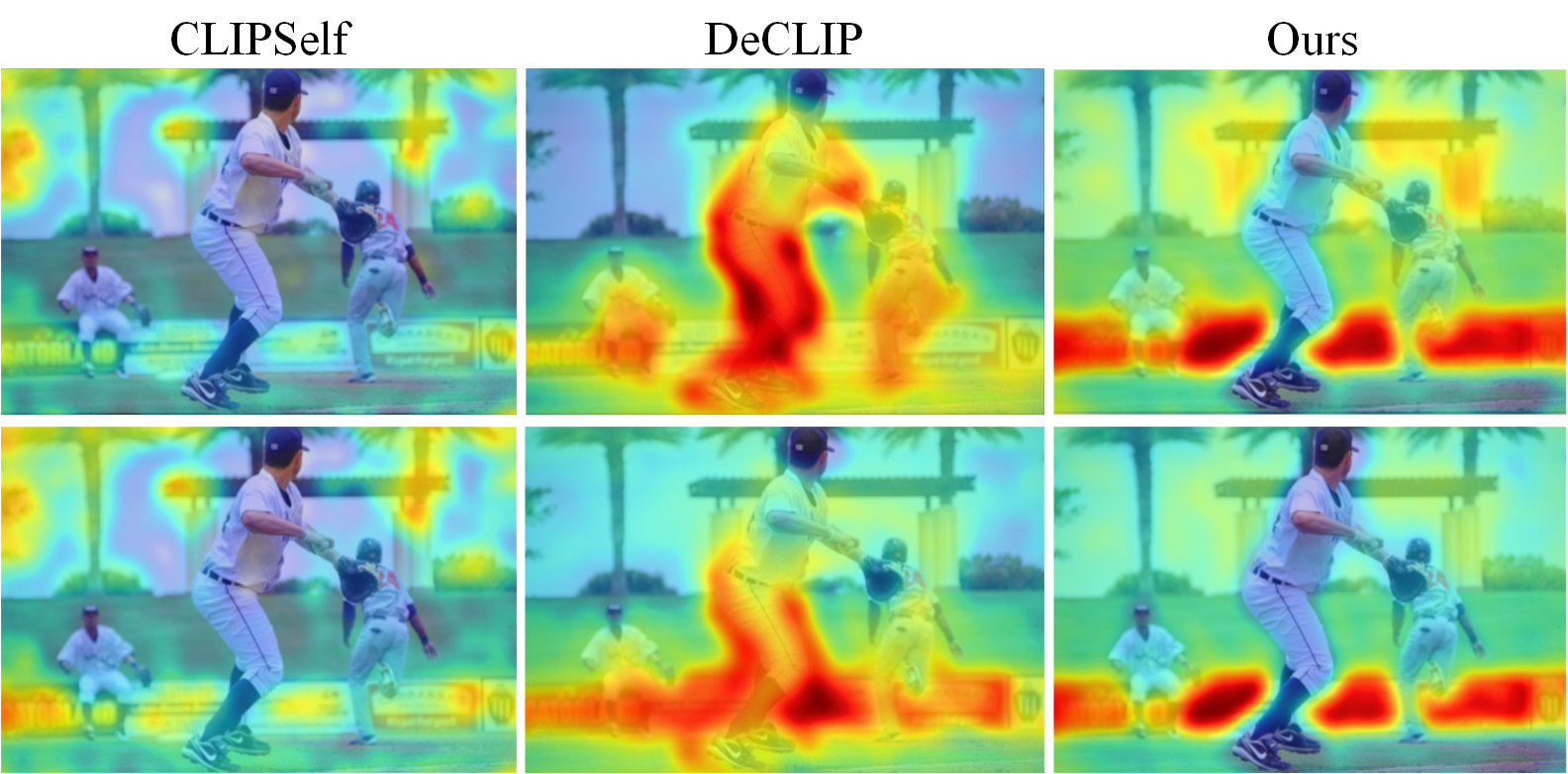}}\\
    \subfloat[Top: \textit{``bowl"} vs. Bottom: \textit{``deep dish"}]{\includegraphics[width=1.0\linewidth]{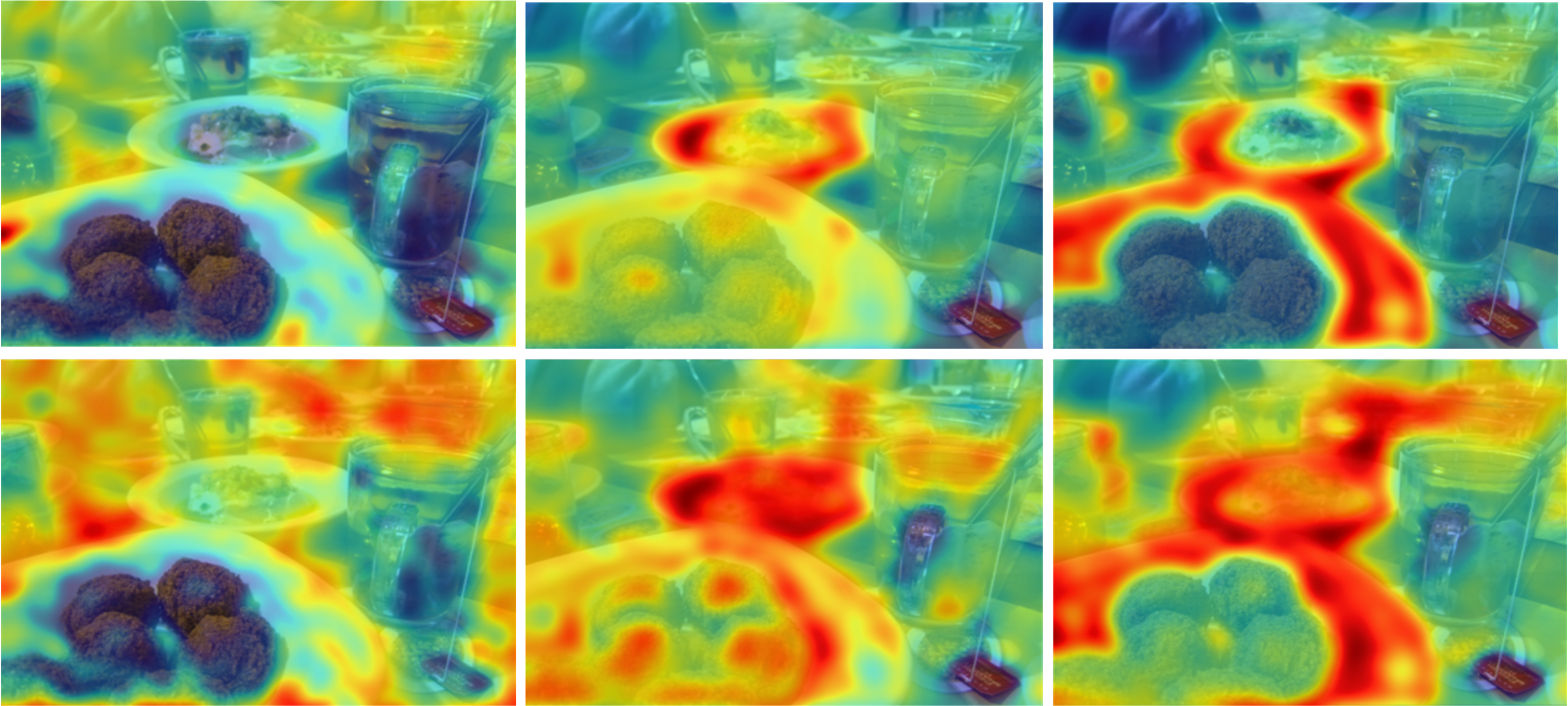}}
    \caption{Qualitative comparison of attention maps under synonymous expressions. (a) \textit{``banner"} vs. \textit{``advertising banner"} and (b) \textit{``bowl"} vs. \textit{``deep dish"}. Compared with CLIPSelf and DeCLIP, SynCLIP maintains more stable and coherent localization across different but semantically equivalent expressions, providing qualitative evidence for its improved robustness to linguistic variations and the effectiveness of synonym-coherent pretraining.}
    \label{fig:syn_attn_vis}
\end{figure}

Table~\ref{tab:synonyms_robust} shows that existing approaches exhibit notable drops when synonyms are used: CLIPSelf decreases by $8.8$ and DeCLIP \cite{wang2025declip} by $9.5$ AP$_{50}^{\text{novel}}$. SynCLIP also experiences a reduction, but the magnitude is smaller with only $4.4$, while base-class performance remains largely unchanged. This suggests that our synonym-coherent pretraining helps reduce the attention shifts introduced by linguistic variations and leads to more stable grounding across semantically equivalent expressions. Figure~\ref{fig:syn_attn_vis} provides complementary qualitative evidence, showing that SynCLIP maintains more coherent attention regions under different synonymous expressions. 

Together, these results align with the grounding inconsistency observed in our motivation and indicate that SynCLIP offers improved robustness to textual variability compared with prior OVDP methods.

\subsection{Ablation Study}

\begin{table}[t]
    \centering
    \begin{tabular}{cc|ccc}
        \toprule
        SSA & SAR & AP$_{50}^{\text{novel}}$ & \gray{AP$_{50}^{\text{base}}$} & \gray{AP$_{50}^{\text{all}}$} \\
        \midrule
        \ding{56} & \ding{56} & 41.0          & \gray{50.9} & \gray{48.3} \\
        \ding{52} & \ding{56} & 41.1          & \gray{51.8} & \gray{48.2} \\
        \ding{56} & \ding{52} & 42.2          & \gray{52.2} & \gray{49.6} \\
        \ding{52} & \ding{52} & \textbf{43.6} & \gray{51.8} & \gray{49.6} \\
        \bottomrule
    \end{tabular}
    \caption{Ablation study of components in SynCLIP with F-ViT on the OV-COCO benchmark. Both SSA and SAR contribute to performance gains, and their combination yields the best novel-category results, demonstrating their complementarity.}
    \label{tab:ablation}
\end{table}

\subsubsection{Component Analysis}
Table \ref{tab:ablation} shows the effectiveness of the main components in SynCLIP with F-ViT through quantitative results on the OV-COCO. Starting from the baseline, we observe an AP$_{50}^{\text{novel}}$ of 41.0 in the first row. Introducing SSA alone yields a marginal improvement to 41.1 AP$_{50}^{\text{novel}}$, confirming that its standalone effect is limited due to the redundant information introduced by semantically enriched expression, which hinders precise spatial attention. In contrast, applying SAR alone significantly boosts AP$_{50}^{\text{novel}}$ to 42.2, highlighting the importance of refining spatial attention with semantic tokens. When combining both components, SynCLIP achieves the best performance with 43.6 AP$_{50}^{\text{novel}}$, demonstrating the complementary benefits of SSA and SAR in enhancing overall robustness and performance.

\begin{figure}[t]
    \centering
    \subfloat[Performance trends]{\includegraphics[width=0.5\linewidth]{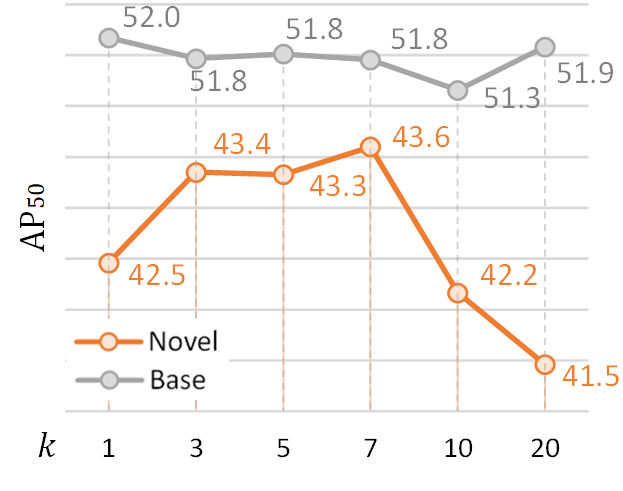}}
    \subfloat[Spatial attention visualization]{\includegraphics[width=0.5\linewidth]{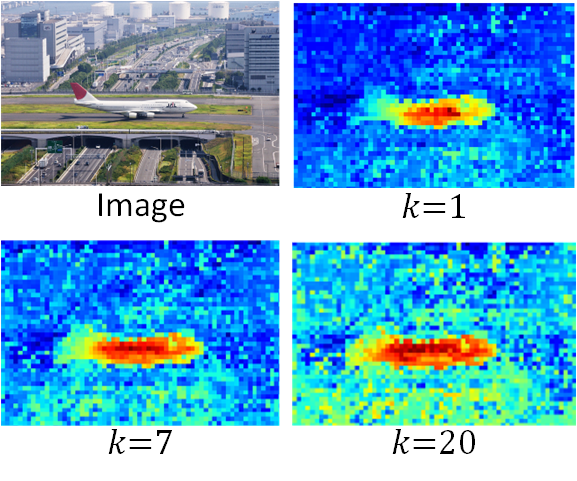}}
    \caption{Effect of the number of semantic tokens, \textit{i.e.}, $k$. (a) shows performance trends on the OV-COCO benchmark with $k$ increasing. (b) shows the qualitative comparison of spatial correlation attention, \textit{i.e.}, $\mathbf{A}_{\text{spa}}$, with various $k$ values.}
    \label{fig:parameter}
\end{figure}

\subsubsection{Effect of Semantic Token Selection}
To analyze the effectiveness of the semantic token selection mechanism in SAR, we investigate how varying the number of semantic tokens, \textit{i.e.}, $k$, influences performance on the OV-COCO benchmark. As shown in Figure~\ref{fig:parameter} (a), increasing $k$ from 1 to 7 progressively enhances AP$_{50}^{\text{novel}}$ from 42.5 to a peak of 43.6. However, further increasing $k$ to 10 and 20 causes the performance to drop to 42.2 and 41.5, respectively. This demonstrates that selecting a moderate number of semantic tokens (\textit{e.g.}, $k=7$) achieves an optimal trade-off between informativeness and noise suppression, as shown in Figure~\ref{fig:parameter} (b). Excessive token introduces irrelevant cues, which degrade spatial grounding accuracy.

\section{Conclusion}
This paper identifies synonym-induced grounding inconsistency as a critical limitation in CLIP-based methods, where synonymous expressions cause divergent spatial attention and hinder robust dense perception. To address this, we propose SynCLIP, which enforces semantic coherence by aligning attention maps across synonymous expressions and enhances spatial precision through semantic token selection and context-aware attention refinement. In addition, we construct SEViC that enriches category-level supervision with synonymous expressions, providing a foundation for pretraining SynCLIP. Extensive evaluations on multiple benchmarks demonstrate that SynCLIP significantly improves grounding consistency and achieves state-of-the-art performance, paving the way for more reliable and robust real-world OVDP applications.
{
    \small
    \bibliographystyle{ieeenat_fullname}
    \bibliography{main}
}

\clearpage
\setcounter{page}{1}
\maketitlesupplementary

\section*{Overview}

This supplementary material provides additional details and results that complement the main paper. Section \ref{sec:sup_sevic} describes the construction pipeline of SEViC. Section \ref{sec:sup_abl} presents additional ablation studies of key components in SynCLIP. Section \ref{sec:sup_vis} provides extended qualitative analyses, and Section \ref{sec:sup_eff} reports the efficiency analysis.

\section{Details of SEViC Construction}
\label{sec:sup_sevic}

\begin{figure*}[t]
    \centering
    \includegraphics[width=1.0\linewidth]{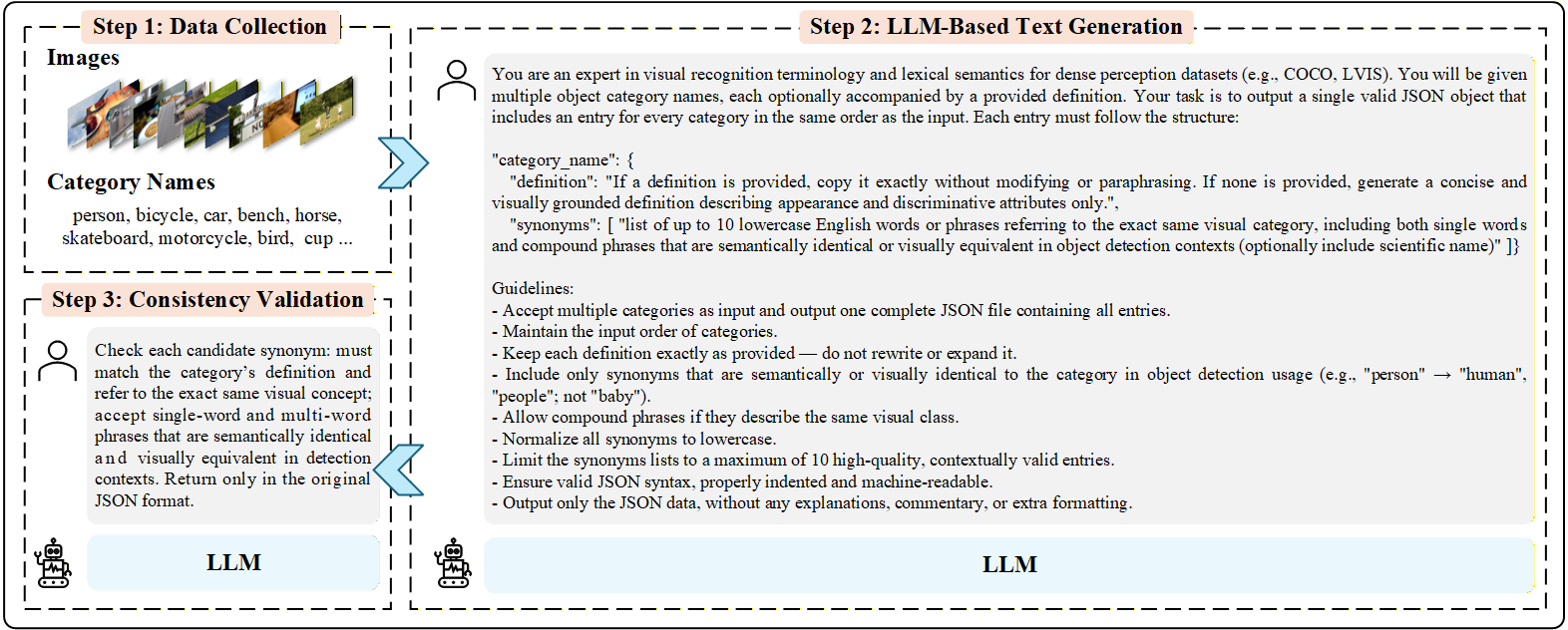}
    \caption{Overview of the SEViC construction pipeline, consisting of (1) Data Collection from COCO and LVIS datasets, which gathers the complete category vocabulary and initial textual resources; (2) LLM-based Text Generation, which expands each category into semantically consistent definitions and synonyms; and (3) Consistency Validation, which filters and verifies all generated expressions through LLM-driven checks to ensure semantic fidelity. These steps jointly produce a semantically coherent, synonym-enriched visual corpus.}
    \label{fig:sevic_build}
\end{figure*}

This section provides a detailed description of the construction pipeline of our synonym-enriched visual corpus (SEViC), complementing the overview in the main paper. As illustrated in Figure~\ref{fig:sevic_build}, the pipeline consists of three major stages, \textit{i.e.}, Data Collection, which gathers the full category vocabulary and initial textual resources; LLM-based Text Generation, which expands each category into semantically consistent expressions; and Consistency Validation, which filters and verifies all generated expressions to ensure semantic fidelity. Together, these stages yield a noise-reduced, semantically aligned corpus that supports synonym-coherent pretraining.

\subsection{Data Collection}

We begin by collecting all category names from COCO2017 \cite{lin2014microsoft} and LVIS \cite{gupta2019lvis}, which share the same image set but differ in granularity and coverage. This provides a unified vocabulary of 1,232 unique object category names and 118,287 images. When available, we also extract their accompanying LVIS-provided definitions and synonyms. These human-curated textual descriptions serve as the initial reference for later LLM-based enrichment. Two meta-categories, \emph{``object"} and \emph{``background"}, are additionally included to represent global semantics.

\subsection{LLM-Based Text Generation}

To enrich category-level linguistic supervision, we employ a large language model (LLM), \textit{i.e.}, DeepSeek \cite{liu2024deepseek}, to generate structured semantic metadata. For each category name, optionally with definitions provided by LVIS, the LLM generates a JSON-formatted semantic entry including: (1) a concise definition that is either preserved verbatim when provided or generated when absent, and (2) a list of synonyms, containing up to ten words or compound phrases, maintaining strict semantic and visual equivalence for dense perception. All generated entries follow a machine-readable JSON format for seamless integration into open-vocabulary pretraining.

\subsection{Consistency Validation}
To eliminate ambiguous or inconsistent expressions introduced by LLMs, we perform a consistency validation using another LLM, \textit{i.e.}, ChatGPT \cite{achiam2023gpt}. Each candidate expression is strictly verified to precisely match the category’s canonical definition and to refer to exactly the same visual object instances as the original label, rejecting overly broad or context-dependent terms. This validation process eliminates noise and ensures that all retained expressions are semantically equivalent, thereby providing high-quality supervision for pretraining.

After the above stages, each collected image is linked to its associated categories along with validated definitions and synonyms, resulting in a high-quality synonym-enriched visual corpus that serves as a foundation for pretraining SynCLIP.

\section{Additional Ablation Studies}
\label{sec:sup_abl}

This section provides additional analyses on two key components of SynCLIP, \textit{i.e.}, the types of semantically enriched textual variants used in SSA for semantic–consistent attention alignment, and the aggregation weights in SAR that balance semantic relevance and spatial precision. These studies further validate the design choices introduced in the main paper and clarify how each component contributes to robust synonym-coherent dense perception.

\subsection{Effect of Textual Variant Types}

\begin{table}[t]
    \centering
    \setlength{\tabcolsep}{1mm}
    \begin{tabular}{cc|cc}
        \toprule
        Synonym & Definition & AP$_{50}^{\text{novel}}$ & \gray{AP$_{50}^{\text{base}}$} \\
        \midrule
        \ding{56} & \ding{56} & 41.1 & \gray{51.7} \\
        \ding{52} & \ding{56} & 42.1 & \gray{51.5} \\
        \ding{56} & \ding{52} & 42.3 & \gray{51.7} \\
        \ding{52} & \ding{52} & \textbf{43.6} & \gray{51.8} \\
        \bottomrule
    \end{tabular}
    \caption{Ablation of textual variant types used in SSA on OV-COCO. The results show that synonyms and definitions provide complementary semantic enrichment, and using both yields the strongest improvement, confirming that SSA benefits from jointly leveraging lexical diversity and explicit semantic grounding.}
    \label{tab:semantic_type}
\end{table}

As shown in Table~\ref{tab:semantic_type}, removing both synonyms and definitions yields the weakest performance, indicating that SSA requires semantically diverse cues to guide attention alignment. Using synonyms or definitions alone already provides clear gains: synonyms introduce lexical variability while retaining the concept, and definitions offer explicit semantic grounding that often captures appearance or functionality. Their combination delivers the highest improvement of 43.6 AP$_{50}^{\text{novel}}$, showing that lexical diversity and semantic specificity contribute complementary benefits. These results confirm that SSA is most effective when aligned to a semantically enriched reference.

\subsection{Effect of Attention Aggregation} 

The SAR module aggregates spatial and semantic attentions via coefficients $\alpha$ and $\beta$. Table~\ref{tab:abl_agg} shows that overly large $\beta$ leads to more concept-driven activations with degraded localization, while overly large $\alpha$ produces spatially precise but semantically weaker maps that generalize poorly to novel categories. Optimal performance on OV-COCO emerges at a balanced configuration with $\alpha=\beta=0.5$, demonstrating that robust dense perception requires both spatial structural cues and semantically discriminative signals. This validates the SAR design and highlights the necessity of harmonizing spatial precision with semantic consistency.

\begin{table}[t]
    \centering
    \begin{tabular}{cc|ccc}
        \toprule
        $\alpha$ & $\beta$ & AP$_{50}^{\text{novel}}$ & \gray{AP$_{50}^{\text{base}}$} & \gray{AP$_{50}^{\text{all}}$} \\
        \midrule
        0.9 & 0.1 & 41.4 & \gray{51.0} & \gray{48.5} \\
        0.8 & 0.2 & 41.8 & \gray{51.3} & \gray{48.8} \\
        0.7 & 0.3 & 41.4 & \gray{51.1} & \gray{48.6} \\
        0.6 & 0.4 & 42.4 & \gray{51.4} & \gray{49.1} \\
        0.5 & 0.5 & \textbf{43.6} & \gray{51.8} & \gray{49.6} \\
        0.4 & 0.6 & 43.1 & \gray{52.1} & \gray{51.8} \\
        0.3 & 0.7 & 43.5 & \gray{52.2} & \gray{49.9} \\
        0.2 & 0.8 & 42.5 & \gray{52.1} & \gray{49.6} \\
        0.1 & 0.9 & 41.7 & \gray{51.7} & \gray{49.1} \\
        \bottomrule
    \end{tabular}
    \caption{Ablation of spatial ($\alpha$) and semantic ($\beta$) attention aggregation weights in SAR on OV-COCO. The results demonstrate that balanced fusion produces the best performance, validating the design of SAR as an effective way to harmonize spatial precision with semantic consistency.}
    \label{tab:abl_agg}
\end{table}

\section{Additional Qualitative Analysis}
\label{sec:sup_vis}

This section presents additional qualitative analyses, including visualizations of attention maps from the SAR module and prediction results on two standard dense perception benchmarks, offering a more comprehensive illustration of the effectiveness and superiority of our method.

\subsection{Visualization of Spatial Attention Refinement}

\begin{figure*}[t]
    \centering
    \includegraphics[width=0.8\linewidth]{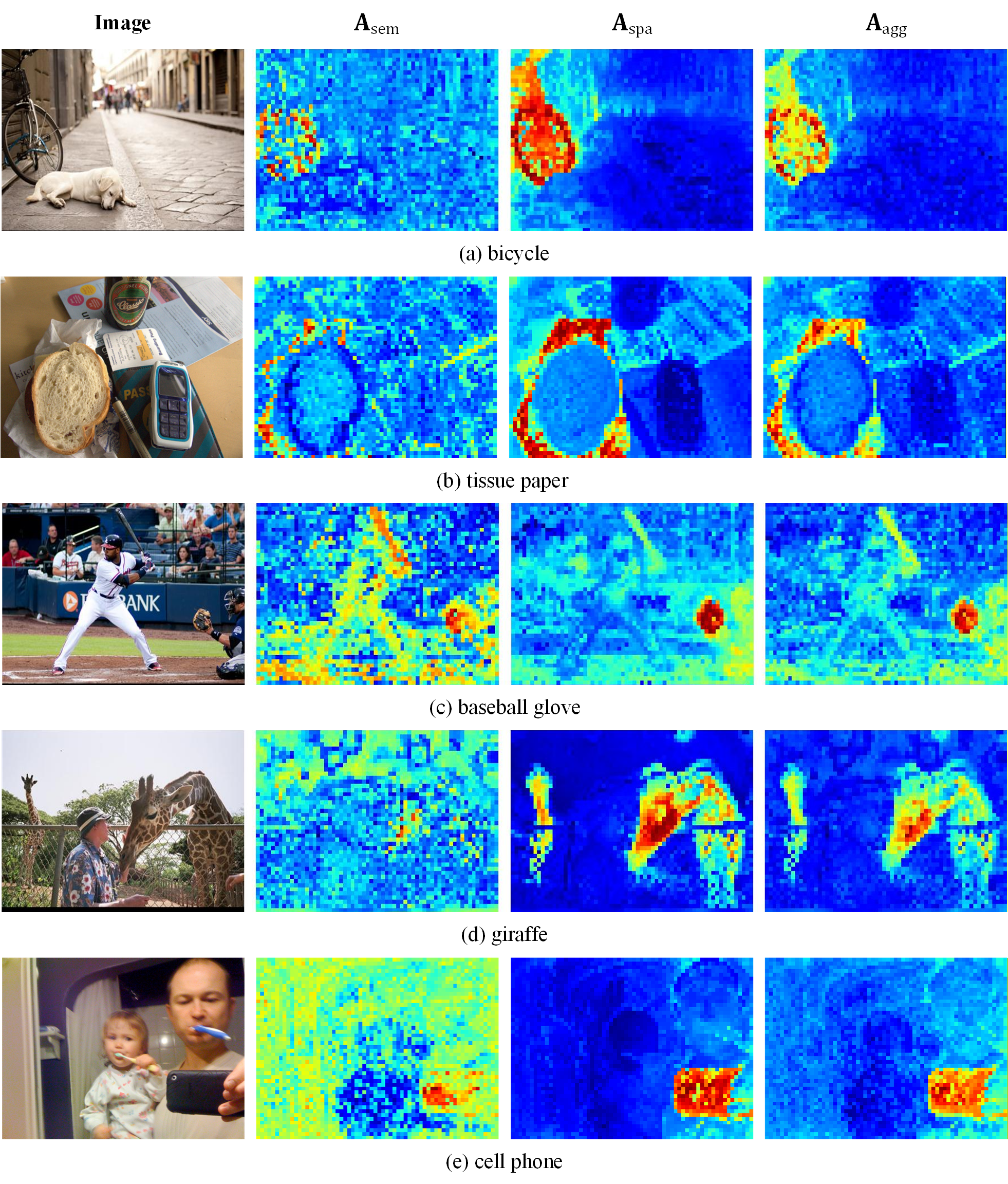}
    \caption{Visualization of five examples from the SAR module with $k{=}7$, showing semantic attention $\mathbf{A}_{\text{sem}}$, spatial correlation attention $\mathbf{A}_{\text{spa}}$, and aggregated attention $\mathbf{A}_{\text{agg}}$. Specifically, $\mathbf{A}_{\text{sem}}$ highlights text-relevant regions but often contains diffuse or noisy activations. $\mathbf{A}_{\text{spa}}$ obtained via semantic token selection provides more accurate localization of target regions, while $\mathbf{A}_{\text{agg}}$, obtained by fusing $\mathbf{A}_{\text{sem}}$ and $\mathbf{A}_{\text{spa}}$, preserves both semantic relevance and text-guided spatial precision.}
    \label{fig:sar_vis}
\end{figure*}

Figure \ref{fig:sar_vis} provides more qualitative examples of attention maps from the SAR module in SynCLIP, including semantic attention $\mathbf{A}_{\text{sem}}$, spatial correlation attention $\mathbf{A}_{\text{spa}}$, and aggregated attention $\mathbf{A}_{\text{agg}}$. Consistent with the analysis in the main paper, $\mathbf{A}_{\text{sem}}$ correctly focuses on concept-relevant areas but may suffer from broad or noisy responses due to lexical variations across enriched expressions. $\mathbf{A}_{\text{spa}}$, computed via semantic token selection and visual foundation model (VFM)-driven spatial contextual reasoning \cite{oquab2023dinov2}, yields concentrated activations around the true object extent. Thus, $\mathbf{A}_{\text{agg}}$ aggregates $\mathbf{A}_{\text{sem}}$ and $\mathbf{A}_{\text{spa}}$, preserving both semantic relevance and text-guided spatial precision. These visualizations highlight the effectiveness of semantic token selection and spatial correlation aggregation in refining text-conditioned spatial attention.

\begin{figure*}[t]
    \centering
    \subfloat[Qualitative results on OV-COCO]{\includegraphics[width=0.7\linewidth]{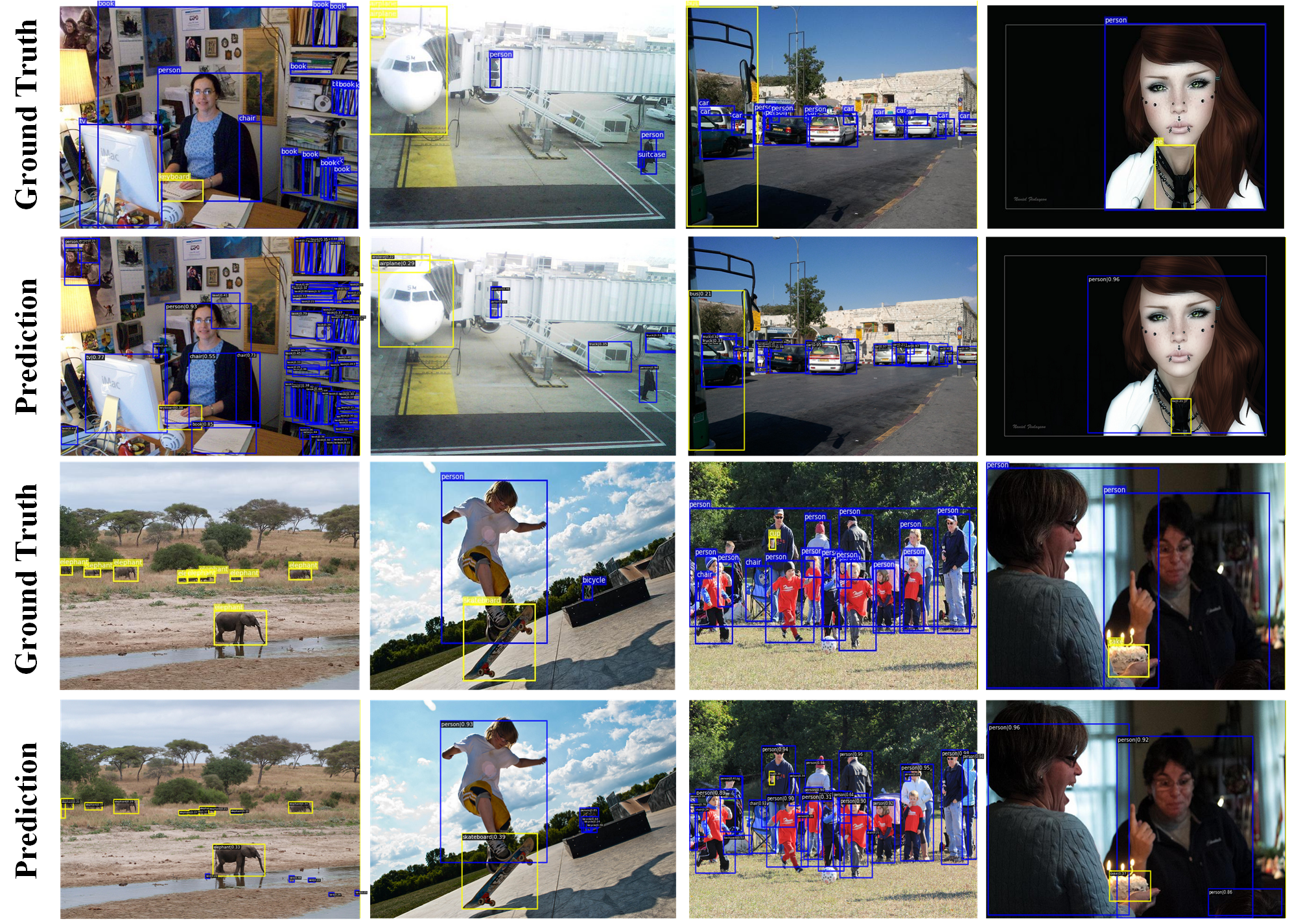}}\\
    \subfloat[Qualitative results on OV-LVIS]{\includegraphics[width=0.7\linewidth]{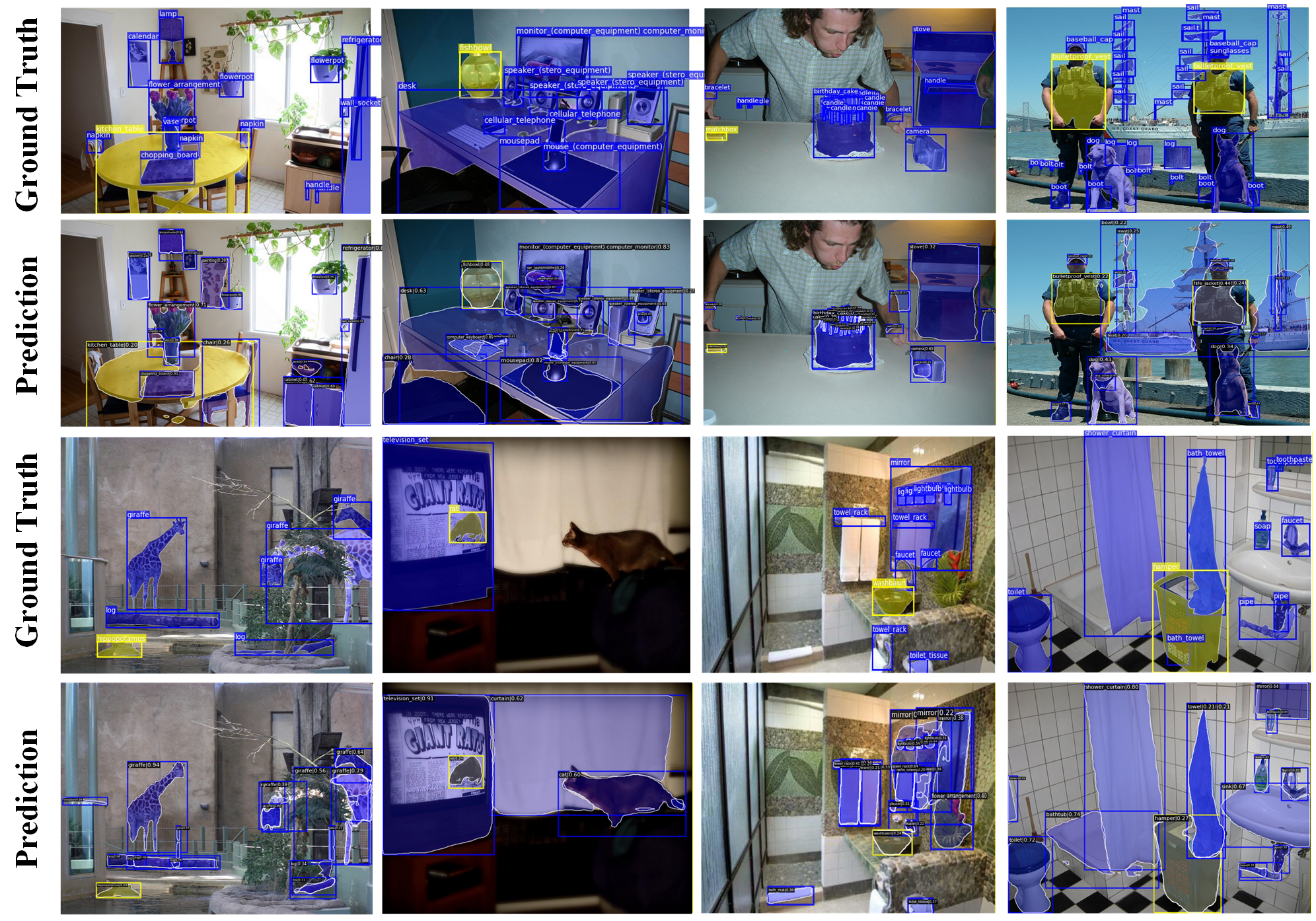}}
    \caption{Qualitative results on (a) OV-COCO and (b) OV-LVIS datasets. Each example includes both ground-truth annotations and predictions from our FViT+SynCLIP model. Yellow boxes denote novel categories, while blue boxes indicate base categories. The model accurately localizes diverse objects and generalizes reliably to novel categories, demonstrating the effectiveness of the proposed method in dense perception.}
    \label{fig:data_res}
\end{figure*}

\subsection{Visualization of Prediction Results}

Figure~\ref{fig:data_res} presents the qualitative results on OV-COCO and OV-LVIS datasets, where yellow and blue boxes correspond to novel and base categories, respectively. It can be observed that our method, \textit{i.e.}, FViT+SynCLIP, successfully detects both base and novel objects with high localization accuracy. These examples demonstrate the model's capability to generalize effectively to novel categories, complementing the quantitative results reported in the main paper.

\section{Efficiency Analysis}
\label{sec:sup_eff}

To provide a comprehensive view of the computational characteristics of SynCLIP, we report the training time, model parameters, FLOPs and AP$_{50}^{\text{novel}}$ on OV-COCO in Table~\ref{tab:time}. All measurements are conducted under the same hardware setup using four NVIDIA A100 GPUs with 40GB memory and an input resolution of 560. Compared with DeCLIP, SynCLIP introduces a modest increase in training time, \textit{i.e.},  one epoch takes 32.5 minutes vs. 19.3 minutes on ViT-B/16 and 144.7 minutes vs. 103.4 minutes on ViT-L/14. This overhead stems from the semantic alignment and attention refinement modules used only during pretraining, which require forwarding frozen auxiliary networks and performing extra attention computations. 

Importantly, these modules are not used during inference. As a result, SynCLIP maintains the same number of parameters and FLOPs as DeCLIP, with no increase in inference-time latency. The observed performance gains therefore stem not from increased model capacity, but from the more structured and semantically consistent pretraining objective, which enhances robustness to downstream dense perception tasks.

\begin{table}[t]
    \centering
    \resizebox{1.0\linewidth}{!}{
        \begin{tabular}{c|c|cccc}
            \toprule
            Method  & Backbone & Time(min) & Params(M) & FLOPs(G) & AP$_{50}^{\text{novel}}$ \\
            \midrule
            DeCLIP  & ViT-B/16 & 19.3      & 86.3      & 33.7     & 41.1 \\
            SynCLIP & ViT-B/16 & 32.5      & 86.3      & 33.7     & 43.6 \\
            \midrule
            DeCLIP  & ViT-L/14 & 103.4     & 304.1     & 155.6    & 46.2 \\
            SynCLIP & ViT-L/14 & 144.7     & 304.1     & 155.6    & 49.8 \\
            \bottomrule
        \end{tabular}
    }
    \caption{Comparison of training time, model size, FLOPs and AP$_{50}^{\text{novel}}$ between SynCLIP and DeCLIP. Although SynCLIP incurs a small pretraining-time overhead, it keeps parameters and FLOPs unchanged at inference while producing substantially higher AP$_{50}^{\text{novel}}$, demonstrating that the benefits arise from synonym-consistent pretraining rather than from larger model capacity or increased inference cost.}
    \label{tab:time}
\end{table}

\end{document}